# Beyond Directed Acyclic Graphs: Causal Zeros and Causal Differential Equations

Sergei V. Kalinin

Department of Materials Science and Engineering, University of Tennessee, Knoxville

## Abstract

Pearl's structural causal model (SCM) framework, built on directed acyclic graphs (DAGs) and the do-calculus, is the dominant formal language for causal reasoning. Yet it carries two structural restrictions: every relationship must be pre-specified as a directed causal edge, and feedback cycles are forbidden. This paper examines two classes of phenomena that strain these restrictions. First, symmetric physical and economic constraints, the ideal gas law being the canonical case, carry no intrinsic causal direction. Direction emerges only under intervention, and which variable is solved for must be specified as part of the intervention. We formalize such constraints as causal zeros within an Extended Causal Model by adding an activation operator, subject to local solvability and graph-admissibility conditions. Second, for the class of finite-propagation state-space systems considered here, we treat apparent instantaneous cycles as artifacts of suppressed time and ground both causal zeros and feedback in Causal Differential Equations (CDEs). In these, the transient regime is a time-unrolled acyclic causal process, and causal zeros arise as the defining functions of attracting equilibrium manifolds; periodic and chaotic attractors define further regimes of the same dynamics, treated through attractor-relative intervention. We give the extended do-calculus, identifiability conditions, counterfactual semantics, and open problems.

## 1. Introduction

Many mechanistic and dynamical physical models are causal by construction, whereas equilibrium and constraint-based descriptions may not carry a unique causal orientation. In a dynamical model, a force produces an acceleration, an applied field induces a polarization, a chemical-potential gradient drives a flux, and a temperature difference drives heat transport. The asymmetry is carried by the mechanism itself, and it is precisely what is required for prediction under intervention: knowing the mechanism, we can say what will happen to the effect if we change the cause, and equally that nothing will happen to the cause if we intervene on the effect. Much of the predictive content of a mechanistic physical theory resides in this asymmetry rather than in the numerical relation alone.

The data-analysis methods now routinely applied to physical systems have the opposite character. Supervised machine learning, including regression, classification, deep networks and their many variants, fits a conditional expectation or distribution relating a set of features to a target. Nothing in the fitted object distinguishes cause from effect. The association it encodes is symmetric: a model of the target given the features can be exchanged for a model of a feature given the target with no statistical penalty, and the designation of which variable is the target is generally supplied by the analyst rather than discovered from the data. This is the familiar observation that correlation is not causation, but in the physical sciences its consequences are significant. A purely correlative model reproduces observed data faithfully and then fails exactly where it is most wanted, namely under intervention - when composition is changed, a field is applied, or a processing parameter is varied. Confounding by an unobserved variable, ubiquitous in multimodal measurement, produces strong and reproducible associations that carry no interventional content.

The discipline that addresses this gap is causal inference and causal discovery, and its modern formal core is Pearl's structural causal model framework together with the associated graphical and algorithmic machinery[1, 2]. What deserves emphasis in the present context is an unusual division of labor. In the physical sciences the causal *directions* are normally supplied by theory via established microscopic mechanism. The causal *formalism* including directed acyclic graphs, structural equations, the do-calculus, and the discovery algorithms built upon them, has by contrast been developed largely within statistics, econometrics, epidemiology and computer science, in domains where the mechanism is unknown and randomization is the gold standard for establishing it. Bringing the two together is therefore not simply a matter of applying an available algorithm to a materials dataset. The two traditions carry different assumptions about what constitutes a variable, what counts as an intervention, and which parts of the model are fixed by prior knowledge rather than learned from data.

The combination is most valuable, and most necessary, in the *observational* physical sciences, where controlled interventions are difficult or impossible and the available information is a large volume of passively acquired measurement. Imaging is the paradigmatic case. Scanning transmission electron microscopy and scanning probe microscopy now routinely yield atomic positions, polarization and strain fields, local chemistry, and spectroscopic response, all measured co-located over extended fields of view and across many thousands of unit cells. Yet intervention at these scales is severely constrained: one cannot randomize the position of a dopant, assign a domain wall to a treatment group, or hold a local strain field fixed while varying the composition around it. The result is exactly the regime for which causal inference machinery was designed - rich, high-dimensional covariation without randomization - and equally the regime in which correlative analysis is least trustworthy, because the number of plausible confounders grows with every additional measurement channel or underpinning mechanisms.

It is notable that the volume of work applying causal discovery in materials and condensed-matter physics remains exceptionally small. The contrast with the Earth and climate sciences is instructive: there, a mature causal-inference subfield has developed over the past decade, with dedicated methods for spatiotemporal and nonlinear data, community benchmarks, and review-level syntheses that connect causal discovery to equation discovery and to the physics of the systems concerned[3-5]. No comparable body of work exists for materials characterization, and the handful of studies that do exist are recent and largely isolated from one another.

The author's own group has pursued this line of research since 2020. The starting point was the observation that machine learning applied to atomically resolved imaging is fundamentally correlative and therefore acutely susceptible to confounding, which motivated a causal analysis of competing atomistic mechanisms in ferroelectrics from high-resolution STEM data[6]. This was developed in parallel with Bayesian model-selection approaches that select between physically motivated descriptions of ferroelectric domain walls using atomically resolved data[7], extended to spatially resolved causal structure in crystalline solids[8], and to the use of predictability itself as a probe of manifest and latent physics in multiferroic Sm-doped $BiFeO_3$[9]. Subsequent work introduced non-Gaussian linear causal models of the LiNGAM family[10], combined with deep kernel learning, to piezoresponse force microscopy, where the non-Gaussianity of the observables supplies the orientation information that covariance alone cannot[11]. The programmatic case for moving from structural description to generative and causal models of condensed matter was set out in Ref. [12], and the argument that scanning probes provide a uniquely favorable platform for causal and interventional learning since the probe can both measure and act is argued in Ref. [13]. Most recently, domain knowledge extracted from the literature by large language models has been used as a prior over graph structure, combined with data-driven discovery on STEM observables to construct directed acyclic graphs linking structural, chemical and polarization degrees of freedom[14].

Work by other groups is similarly sparse but growing. Causal relations derived from first-principles data have been used to identify the structural modes governing cation ordering in double perovskites[15], and LiNGAM-based analysis has been applied to the factors controlling lithium ionic conductivity in olivine-type materials[16]. In catalysis, causal inference has been proposed as the route from correlative process-structure models to actionable mechanistic ones[17], and broader perspectives have argued for the integration of explainable, interpretable and causal modeling into materials research as a whole[18]. Taken together, this is a small literature by the standards of either parent field, and it is dominated by the application of existing algorithms rather than by the adaptation of the formalism to the specific structure of physical problems.

In the course of this exploration we came to notice a recurring obstacle that is not algorithmic but conceptual: certain elementary physical constructs do not map onto the Pearl formalism at all. The clearest of these, and the subject of this manuscript, is what we will call a *causal zero*. Consider the ideal gas law, $PV = nRT$. It is a perfectly good physical relation, it supports prediction, and it certainly supports intervention - compress the gas and the pressure responds, heat it at fixed volume and the pressure responds again. But it contains no arrow. Nothing in the physics designates pressure as the cause of volume or volume as the cause of pressure, and which variable becomes the effect is determined not by the law but by the experiment performed on it. The structural causal model framework cannot accommodate this directly, because it requires that the direction be committed before any intervention is specified: the graph must already contain or exclude the edge before $do(X = x)$ can be evaluated. Equilibrium relations, conservation laws, constitutive relations, and accounting identities all share this structure, and they are pervasive in physics.

The natural place to look for a remedy is the literature on cyclic causal models, and in particular the sustained program of Mooij and co-workers, which has produced the most complete apparatus available for equilibrium and feedback systems, including constraint-based generalizations of structural causal models, foundations for cycles and latent variables, a generalized causal calculus, and explicit bridges from differential equations to causal models[19-23]. This work is mathematically powerful and we rely on it heavily in what follows. The present paper adopts a narrower modeling stance: for finite-propagation, well-posed physical state-space models without algebraic loops, feedback is mediated through dynamics rather than through literally simultaneous influence. A cycle drawn among variables at one instant can then be understood as an artifact of suppressing time, and restoring time provides an acyclic unrolling. Cyclic SCMs and causal SDEs remain legitimate at coarser, equilibrium, or sample-path levels; recent work establishes graphical Markov properties and do-calculus for causal SDEs under explicit solvability conditions[24]. The symmetric relations that survive the dynamical refinement are constraints rather than directed cycles, and they call for a different treatment.

This manuscript summarizes an attempt to give that treatment. We propose the causal zero as a first-class object: a symmetric, lawlike constraint that carries no baseline orientation and is represented as a hyperedge rather than a directed edge. We show how causal zeros arise in equilibrium physics, and specifically how thermodynamic equilibration destroys causal orientation. During relaxation the process is directed and causal - the fast variable is driven by the slow ones - but once equilibrium is reached, the surviving algebraic relation among the state variables retains no record of which variable was driven. The arrow is not hidden by the equilibrium description; it is erased by it. This connects the construction directly to the theory of stationary states of differential equations: a causal zero is the defining function of an attracting equilibrium manifold of an underlying dynamical system, and the familiar static causal model is recovered as the special case in which that manifold degenerates to an isolated point.

Finally, we introduce an activation operator that resolves a causal zero into ordinary causal structure. Because a single scalar constraint can determine exactly one variable given the others, an intervention on a causal zero must specify not only which variable is externally fixed (the driver), but also which variable the constraint is solved for (the target). Supplied with both, and subject to local solvability and graph-admissibility conditions, the activation operator compiles the undirected hyperedge into a conventional directed acyclic graph together with an explicit structural equation, after which the standard do-calculus applies unchanged. The requirement that the target be named explicitly is not a technical convenience; it is the same lesson taught by labeled equilibrium equations[20], and omitting it renders the operation ill-posed whenever the constraint involves more than two variables. The remainder of the paper develops these ideas: Section 2 reviews the Pearl framework, Sections 3 and 4 examine the two structural limitations, Section 5 formalizes causal zeros, and Section 6 grounds the construction in causal differential equations.

## 2. Pearl's Structural Causal Framework

### 2.1 Motivation and Historical Context

The problem of distinguishing causation from correlation is among the oldest in scientific epistemology. Hume's skepticism about causal necessity, Mill's methods of induction, and twentieth-century debates in philosophy of science all grappled with what it means for one event to cause another. Statistics, for most of the twentieth century, deliberately avoided causal language: the dominant frequentist and Bayesian frameworks concerned themselves with associations in data, not with the mechanisms that generated them.

The modern formal theory of causation owes its most systematic development to Judea Pearl[1, 25], whose work through the 1990s and culminating in the monograph Causality provided the first mathematically complete language for causal reasoning, intervention, and counterfactual inference. Pearl's framework synthesizes graphical models, structural equations, and probabilistic logic into a unified do-calculus capable of answering questions purely associational statistics cannot: what would happen if we intervened, and what would have happened had circumstances differed.

### 2.2 Structural Causal Models

A Structural Causal Model (SCM) is a tuple $M = (V, U, F, P_U)$, where $V = \{X_1, \dots, X_n\}$ is a set of endogenous (observed) variables, $U = \{U_1, \dots, U_n\}$ is a set of exogenous (background) variables with joint distribution $P_U$, and $F = \{f_1, \dots, f_n\}$ is a set of structural functions:

$$X_i \coloneqq f_i(\mathrm{PA}_i, U_i), \quad i = 1, \dots, n$$

where $\mathrm{PA}_i$ denotes the parents of $X_i$ in the causal graph G, the directed acyclic graph induced by F. Each structural equation encodes a generative mechanism: the assignment operator ':=' signifies that $X_i$ is determined by its parents and noise, not merely correlated with them. As Bochman and Lifschitz[26] emphasize, this asymmetry is carried by the assignment syntax itself, independently of the numerical equality the assignment happens to satisfy - a point we return to in Section 3.

### 2.3 The *do*-Calculus and Graph Surgery

The central innovation of the framework is the *do*-operator. The intervention $do(X_i = x)$ represents an external manipulation setting $X_i$ to $x$ regardless of its natural causes. It is implemented as graph surgery: all incoming edges to $X_i$ are severed, yielding the mutilated graph in which the structural equation of $X_i$ is replaced by the constant assignment $X_i \coloneqq x$. The interventional distribution is:

$$P(y \mid do(x)) = P_{M_x}(y)$$

where $M_x$ is the modified model. This answers a question passive observation cannot: in a population where $X$ is forced to $x$ by intervention, what is the probability of $P(Y = y)$? It differs fundamentally from the conditional distribution $P(y \mid x)$.

### 2.4 The Three Rules of do-Calculus

The do-calculus provides three rules for transforming interventional expressions. Below, the overline denotes deletion of edges into a variable set and the underline denotes deletion of edges out of a variable set.

Rule 1: Insertion/deletion of observations:

$$P(y \mid do(x), z, w) = P(y \mid do(x), w) \text{ if } (Y \perp\!\!\!\perp Z \mid X, W)_{G_{\overline{X}}}$$

Rule 2: Action/observation exchange (here the graph also has edges out of Z removed):

$$P(y \mid do(x), do(z), w) = P(y \mid do(x), z, w) \text{ if } (Y \perp\!\!\!\perp Z \mid X, W)_{G_{\overline{X},\underline{Z}}}$$

Rule 3: Insertion/deletion of actions:

$$P(y \mid do(x), do(z), w) = P(y \mid do(x), w) \text{ if } (Y \perp\!\!\!\perp Z \mid X, W)_{G_{\overline{X},\overline{Z(W)}}}$$

These three rules, together with standard probability calculus, are sound and complete[27, 28] for identifying causal effects in DAG-structured SCMs: if they cannot reduce an interventional query to an observational one, no method can.

### 2.5 The Ladder of Causation

Pearl[29] organizes causal reasoning into three levels: association (statistical correlations, $P(y \mid x)$); intervention (effects of actions, $P(y \mid do(x))$); and counterfactuals (what would have happened under alternative circumstances). The hierarchy makes explicit that purely associational learning operates only at the first level and cannot, by itself, answer interventional or counterfactual questions.

### 2.6 Key Derived Results

Perlean formalism allows a broad set of the causal inference algorithm starting from the celebrated front-door and backdoor adjustments. A set Z satisfies the backdoor criterion relative to (X, Y) if no node in Z is a descendant of X and Z blocks every path from X to Y with an arrow into X. Then:

$$P(y \mid do(x)) = \sum_{z} P(y \mid x, z)\, P(z)$$

permitting computation of an interventional distribution from observational data by adjusting for observed confounders.

When unmeasured confounding links X and Y but a mediator M satisfies suitable conditions, the effect is still identified via the front-door formula of Pearl[25]:

$$P(y \mid do(x)) = \sum_{m} P(m \mid x) \sum_{x'} P(y \mid x', m)\, P(x')$$

which achieves identification even under unobserved confounding, given the mediating structure.

## 3. Limitation I: Non-Causal Symmetric Relationships

### 3.1 The Asymmetry Assumption

The SCM framework rests on a fundamental asymmetry assumption: every relationship is a directed causal edge, an unobserved common cause, or nothing. The DAG encodes this structurally - arrows have direction, and direction represents the flow of causal influence. For many phenomena this is natural: smoking causes lung cancer, not the reverse; altitude lowers air pressure, not the reverse. For such phenomena, causal direction is pre-given by physical law, temporal order, or experimental design.

### 3.2 Symmetric Physical Constraints

A large and important class of relationships, however, is intrinsically symmetric. The ideal gas law is the canonical example:

$$PV = nRT$$

relating pressure P, volume V, amount n, and temperature T. It is an equilibrium constraint: at equilibrium these quantities co-determine one another, and the physics supplies no arrow from P to V or V to P. Which variable becomes the effect is fixed by the intervention, not by the equation: compress the gas (fix V, hold T) and P is determined; heat it (fix T) and the constraint relates P and V along an isotherm. Crucially, one intervention pins one variable - a single scalar equation has exactly one degree of freedom to close - so a complete specification of 'intervene here' must also say 'and solve for there'. This observation, trivial as it seems, is what forces the target-labeling in Section 5.

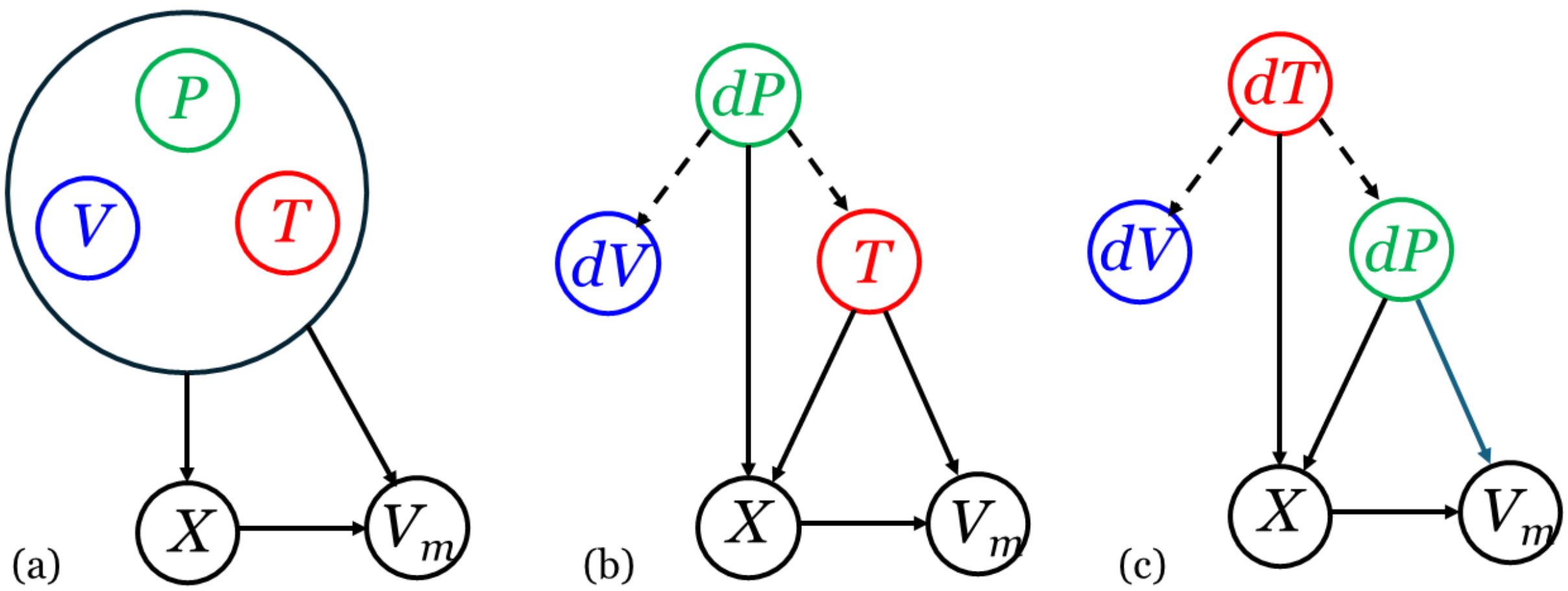

**Figure 1.** The causal zero concept. (a) Shown is the relationship between a gas phase characterized by pressure P, volume V, and temperature T, in equilibrium with a solid characterized by the concentration of species X and molar volume Vm. The values (P, V, T) affect X through electrochemical equilibrium (equality of electrochemical potentials), and Vm through both X and compressibility. However, (P, V, T) do not form a DAG because their mutual connection is a symmetric equilibrium constraint. Upon intervention on (b) pressure or (c) temperature, the (P, V, T) hypernode is activated and forms a classical causal DAG. In panels (b) and (c), dP, dV, and dT denote controlled perturbations; dashed links indicate variables constrained by the intervention protocol.

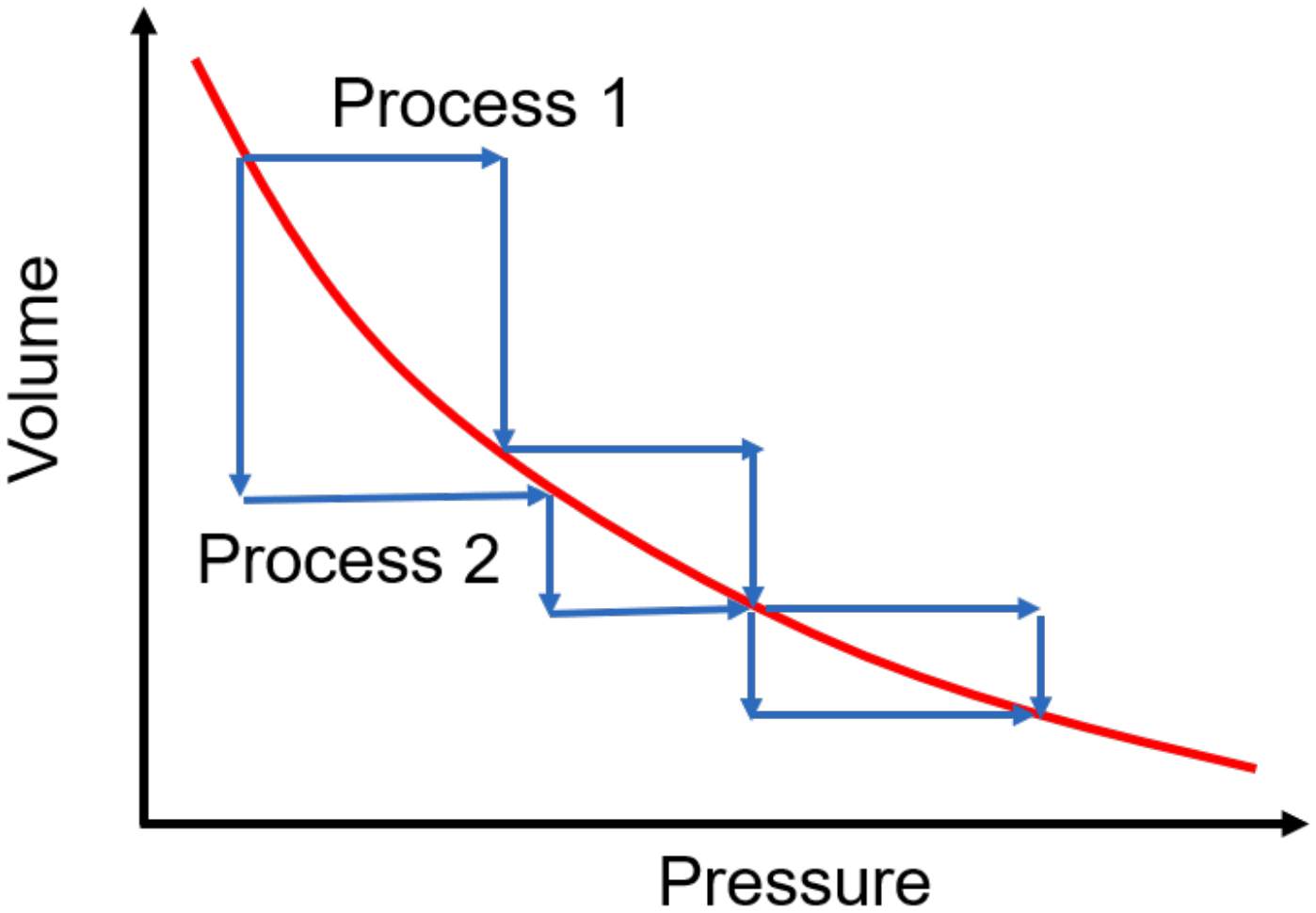

**Figure 2.** Compression of the ideal gas driven by pressure change at constant volume or volume change at constant pressure are causal process. However, in the limit of the infinitely small change (isothermic process) the system becomes non-causal.

### 3.3 Further Examples

The pattern recurs across sciences. In economics a supply-demand equilibrium jointly fixes price and quantity; a supply shock makes quantity drive price, whereas a demand shock can produce the reverse orientation. In circuits, Ohm's law may be solved for voltage or current depending on whether a current source or voltage source is imposed. In chemistry, an equilibrium constant constrains concentrations, and which species is driven depends on what is added or held fixed. The relation F = ma is instructive for a different reason: solving it for the force required to attain a desired acceleration changes the computational target in an inverse-design problem, but does not by itself reverse the physical causal direction. Equilibrium and constitutive constraints therefore provide the cleaner examples of orientation emerging from an intervention protocol. None of these constraints carries a unique arrow by itself; each acquires one only when an intervention also selects a solved-for quantity.

### 3.4 The Fundamental Gap

Standard do-calculus has no vocabulary for such relationships: before any $do(X = x)$ is applied, the DAG must already contain or exclude the arrow $X \rightarrow Y$. Imposing a direction in advance misrepresents the relationship and can license interventions the physics forbids, Figure 1, 2. As Bochman and Lifschitz[26] note, $PV = nRT$ is not, as written, the structural assignment $P :=$ $nRT/V$; the equality is symmetric while the assignment is not. The precise gap we address is the absence of a device that (i) stores a lawlike relation without a committed direction and (ii) compiles it into a specific directed mechanism at intervention time, given a driver and a target.

## 4. Limitation II: Cycles and the Prior Work on Both

### 4.1 The DAG Acyclicity Requirement

The SCM framework requires G to be acyclic, and this requirement is load-bearing: $d$-separation is defined on DAGs; the Markov factorization assumes acyclicity; the topological order used to solve the structural equations requires it; and graph surgery is well-defined only when no path leads from an intervened variable back to itself. The standard discovery algorithms built on the framework such as PC, FCI[2], and descendants likewise presuppose acyclicity.

### 4.2 Why Cycles Arise

Feedback is nonetheless ubiquitous. Gene-regulatory networks show mutual activation and repression; Lotka–Volterra predator–prey dynamics couple two populations in mutual dependence; the IS–LM model couples income and interest through simultaneous equations; recurrent neural circuits carry dense feedback; a thermostat measures temperature, drives a heater, and thereby changes temperature. In each case the loop is the mechanism, not a modeling artefact.

### 4.3 Prior Approaches to Cyclic Causal Models

#### *4.3.1 Equilibrium and fixed-point approaches*

Simon[30] proposed that the causal order of a simultaneous-equation system is fixed by the minimal self-contained subsets of equations. Iwasaki and Simon[31] extended causal ordering to physical systems, treating the equations as an acausal formal description from which an ordering is inferred; tellingly, the same equilibrium law yields different orderings depending on which variables are held exogenous. Spirtes[32], and Strotz and Wold[33], interpreted cyclic structural systems by requiring a unique stable fixed point of the self-map $X = f(X, U)$; Mooij and collaborators[34, 35] later systematized this interpretation. Under unique solvability, interventional semantics is partially recovered by surgery followed by re-solving.

#### *4.3.2 Mooij's program: labeled equilibrium equations and causal constraints models*

A sustained and directly relevant program, due to Mooij and collaborators, treats equilibrium and dynamical causality as primary. Mooij, Janzing and Schölkopf[20] show that the equilibria of a first-order ODE system can be described by a deterministic SCM, but only once the equilibrium equations are labeled by the variable for which each equation is solved. Their central conclusion is exactly ours: an unlabeled set of equilibrium equations loses intervention information because it does not record which equation is replaced under a given intervention. We adopt their labeling discipline directly; it is the reason the activation operator of Section 5 carries an explicit target.

Blom, Bongers and Mooij[19] sharpen this into Causal Constraints Models (CCMs), arguing that SCMs are not flexible enough to represent dynamical systems at equilibrium. A CCM constraint carries an activation set specifying the intervention targets under which it remains in force, and they use the ideal gas law as their running example: representing it as an SCM requires duplicating the law and can admit interventions that violate it, whereas a CCM keeps it as a single constraint that simply has no solution under inconsistent targets. This is the closest existing mechanism to a causal zero; the distinction we draw in Section 5.8 is that a CCM activation set is an admissibility filter, whereas our activation operator additionally compiles the constraint into a directed mechanism.

On the dynamical side, Bongers, Blom and Mooij[36] give Structural Dynamical Causal Models via random differential equations, with conditions under which equilibrating sample paths correspond to SCM solutions; Rubenstein, Bongers, Schölkopf and Mooij[37] generalize to dynamic SCMs representing asymptotic behavior under time-dependent interventions; and Boeken and Mooij[23] refine dynamic SCMs for stochastic differential equations, adding time-splitting and subsampling. Ryan and Dablander[38] develop Equilibrium Causal Models as causal abstractions of dynamical systems for cross-sectional data, emphasizing long-term intervention effects and conditions for estimation. This sequence is the most developed prior treatment of differential dynamics and their equilibria as causal objects, and it directly anticipates the relationship we draw between Sections 5 and 6.

The same group also pursues the opposite reduction, introducing cycles as legitimate static objects through modular SCMs and sigma-separation[22, 39], and PAG-based discovery for cyclic systems[40]. More recently, Boeken and Mooij[24] establish sigma-separation and do-calculus at the sample-path level for causal SDEs under solvability conditions, and prove a stronger d-separation property for additive-noise SDEs even when the system is cyclic. That line generalizes causal graphs to accommodate cycles. Here, we adopt the more restricted physical modeling stance that finite-propagation, well-posed state dynamics should be represented explicitly in time; we do not claim that cyclic SCM or SDE representations are mathematically illegitimate.

#### *4.3.3 Acyclic directed mixed graphs*

Richardson and Spirtes[41] introduced acyclic directed mixed graphs and maximal ancestral graphs, admitting bidirected edges for unobserved common causes. Crucially, a bidirected edge encodes confounding, i.e. a hidden U with $U \rightarrow X$ and $U \rightarrow Y$, not the mutual determination of a symmetric law. Richardson[42] further showed that m-separation characterizes the independences of such models.

#### *4.3.4 Cyclic SCMs, time-series, and discovery*

Bongers, Forré, Peters and Mooij[21] give the most thorough theory of cyclic SCMs, characterizing when interventional and counterfactual distributions are well defined via solvability, uniqueness and stability, and noting that observed cycles can reflect a DAG at a finer level of description. Dynamic Bayesian networks, from Dean and Kanazawa[43] through Murphy[44], unroll cycles across time so that a loop becomes a lagged dependence and the DAG is restored by temporal indexing; Granger[45] causality operationalizes temporal precedence in prediction.

In econometrics, simultaneous-equation models since Haavelmo[46] and Koopmans[47] represent mutual dependence, and their identification theory is a direct precursor to Pearl's; his contribution was the nonparametric, graphical, and complete calculus. On discovery for cyclic systems, Hyttinen, Eberhardt and Hoyer[48] characterize linear cyclic models with latent variables, and the answer-set-programming approach of Hyttinen, Eberhardt and Järvisalo[49] resolves conflicting constraints; Mooij and Heskes[35] show identifiability of linear cyclic models from equilibrium data, paralleling LiNGAM[10] results for the acyclic case.

A parallel continuous-time tradition including Sokol and Hansen[50] on the causal interpretation of stochastic differential equations, Eichler and Didelez[51] and Eichler[52] on graphical time-series and interventions, White, Chalak and Lu[53] linking Granger and Pearl via settable systems, and Røysland, Ryalen, Nygård and Didelez[54] on continuous-time event-history analysis with local independence develops causal semantics directly in time rather than on static graphs.

### 4.4 The Residual Problem

Despite this substantial body of work, a specific gap remains. Existing frameworks either (a) require a committed direction in advance, or (b), as in the case of CCMs, represent a symmetric constraint faithfully but treat its activation as an admissibility filter rather than as an orientation operation that yields a predicted value for the solved-for variable. No existing framework supplies an explicit operator that compiles an undirected constraint into a specific directed mechanism, given a driver and a target, while remaining inside Pearl's calculus after activation. Nor does any single object represent both transient dynamics and equilibrium constraints as regimes of one another. These are the two gaps this paper addresses; the first is a genuine refinement of CCMs rather than a claim of unoccupied territory.

**Table 1.** Comparison of causal frameworks relevant to symmetric constraints and dynamical feedback.

| Framework | Primary object | Symmetric constraints | Dynamics / cycles | Intervention and target semantics |
|---|---|---|---|---|
| Pearl SCM | Directed structural assignments | Not first-class | DAG; no directed cycles | do replaces a preassigned target equation. |
| Cyclic SCM / sigma-separation | Directed equations with feedback | Not primary | Cycles allowed under solvability conditions | Surgery acts on equations whose targets are already assigned. |
| Labeled equilibrium SCM | Equilibrium equations labeled by solved-for variable | Only after labeling | Equilibrium abstraction of ODEs | Intervention replaces the equation carrying the relevant target label. |
| Causal Constraints Model (CCM) | Constraints with activation sets | Yes | Represents equilibrium and feedback summaries | Intervention selects which constraints remain active; no required target-indexed compilation into an SCM. |
| Constrained causal models (Beckers-Halpern-Hitchcock) | SCM plus admissible-setting constraints | Yes, especially identities and setting constraints | Dynamics not primary | Disconnection operation changes how a variable participates in a causal equation. |
| SDCM / DSCM / causal SDE | Random differential equations or time-indexed processes | Can arise through asymptotic behavior | Time dependence and cycles are explicit | Time-dependent or path-level interventions; solvability determines semantics. |

| Framework | Primary object | Symmetric constraints | Dynamics / cycles | Intervention and target semantics |
|---|---|---|---|---|
| Equilibrium Causal Models (Ryan-Dablander) | Equilibrium causal abstraction of a dynamical system | Equilibrium relations are central | Long-run equilibrium rather than transients | Targets long-term intervention effects from cross-sectional equilibrium data. |
| Proposed ECM / causal zero | DAG plus undirected constraint hyperedges | Yes, as first-class causal zeros | Grounded in CDE solution regimes | Activation names driver and solved-for target, then compiles the constraint into an assignment subject to admissibility. |

## 5. Causal Zeros: Extending the Formalism

### 5.1 Conceptual Foundation

A causal zero is a symmetric, non-directional constraint among variables that (i) holds as a functional equation at equilibrium, (ii) carries no intrinsic causal direction, and (iii) becomes a specific directed mechanism admitting a well-defined interventional distribution once an intervention names both a driver and a solved-for target. The name marks the zero point of causal asymmetry: latent causal content with no expressed direction. The constraint

$$\varphi_z(X_{i_1}, \dots, X_{i_k}) \;=\; 0$$

is the zero-level object; a directed mechanism emerges from it only under activation. Because the constraint is a single scalar equation, it can determine exactly one variable given the others - so, unlike a vertex of a DAG, a causal zero does not by itself say what causes what, and cannot be activated without a target.

### 5.2 Formal Definitions

***Definition 5.1 (Causal zero)***

A causal zero is a tuple

$$z \;=\; (V_z, \varphi_z, \Omega_z, \tau_z)$$

where $V_z$ are the constraint variables; $\varphi_z : \mathbb{R}^k \to \mathbb{R}$ is a smooth scalar constraint whose zero set is a smooth $(k-1)$-dimensional manifold; $\Omega_z \subseteq V_z$ is the set of admissible drivers (variables that may be externally fixed); and $\tau_z \subseteq V_z$ is the set of admissible targets (variables the constraint may be solved for). A variable is an admissible target iff $\varphi_z$ is smoothly invertible in it on the zero set — i.e. its partial derivative is nonzero there, so the implicit function theorem applies. For the ideal gas law every variable is both an admissible driver and an admissible target; for a conservation law some variables may be neither.

***Definition 5.2 (Extended Causal Model)***

An Extended Causal Model (ECM) is a tuple

$$\mathcal{M} = (V, U, E_C, E_Z, F, P_U)$$

where V is the set of variables, U the exogenous variables with distribution $P_U$, $E_C$ a set of directed edges forming a DAG $G_C$, $E_Z = \{z_1, \dots, z_m\}$ a set of causal zeros, and F the structural functions attached to $E_C$. The full object is a hypergraph: directed edges alongside undirected constraint hyperedges.

### 5.3 The Activation Operator

***Definition 5.3 (Activating intervention)***

Let $z = (V_z, \varphi_z, \Omega_z, \tau_z)$ be a causal zero, let $X_d \in \Omega_z$ be the driver, and let $X_t \in \tau_z$, with $t \neq d$, be the target. The activating intervention

$$\mathrm{do}_z(X_d = x;\, X_t): \ z \mapsto \left(X_{V_z \setminus \{t\}} \to X_t\right), \ \ X_t := {\varphi_z}^{-1}\left(x, X_{V_z \setminus \{d,t\}}\right)$$

(i) removes z from the constraint set; (ii) fixes the driver at the intervened value; (iii) adds directed edges into the single target from every other constraint variable, $(V_z \setminus \{t\}) \to X_t$; and (iv) installs the structural equation

$$X_t := g_{z,t}\left(X_{V_z \setminus \{t\}}\right) = {\varphi_z}^{-1}\left(X_{V_z \setminus \{t\}}\right), \quad t \in \tau_z$$

obtained by solving the constraint for the target given all the other constraint variables (the driver at its fixed value, the remaining variables as parents). The remaining constraint variables other than $X_d$ and $X_t$ are not pinned by this activation; they retain whatever status they have elsewhere in the model (observed inputs or targets of separate interventions). This is exactly the labeling approach of Mooij, Janzing and Schölkopf[20]: a causal zero is inert until an equation is assigned to a target. Activating with a driver alone, as an earlier version of this work did, is ill-posed whenever $k > 2$ because one scalar equation cannot pin $k - 1$ variables at once. The resulting model is an ordinary DAG-based SCM only when the activation is admissible in the sense below; otherwise the activation may produce a cyclic SCM and requires the corresponding cyclic semantics.

*Definition 5.4 (Activation admissibility)*

An activation $\mathrm{do}_z(X_d = x; X_t)$ is admissible relative to an ECM when: (A1) local solvability holds, $\partial\phi_z / \partial X_t \neq 0$ in the neighborhood of interest; (A2) the imposed driver value and retained constraint variables lie in the local domain of the resolution function and are jointly compatible with the remaining model; and (A3) graph admissibility holds: for every newly added parent $X_j$ among the constraint variables other than $X_t$, the pre-activation directed graph contains no directed path $X_t \rightsquigarrow X_j$. Conditions (A1) and (A2) make the resolved assignment well defined. Condition (A3) ensures that adding $X_j \to X_t$ does not create a directed cycle. If (A3) fails but the resulting cyclic system remains uniquely solvable, activation may still be meaningful, but it no longer compiles the ECM into an ordinary DAG-based SCM and Rules 1-3 must be replaced by an appropriate cyclic calculus.

*Rule 0 (Activation)*

For an ECM with causal zero z, driver $X_d$ in $\Omega_z$ and target $X_t$ in τz, and for an activation satisfying Definition 5.4:

$$P(y \mid \mathrm{do}_z(X_d = x;\, X_t), w) = P(y \mid do(x), w) \mid_{G_{d\to t}{}^z}$$

where the graph on the right is the causal DAG augmented with the admissible edges into the target and the structural equation above. After an admissible activation, the query is an ordinary interventional query and the standard calculus applies. Where several targets are of interest, one activates separately labeled copies of the constraint, one per target, following the labeled-equation discipline rather than pretending that a single scalar equation orients several targets at once.

**5.4 Interventional Pushforward Through an Activated Constraint**

Suppose a causal zero has driver $X$, target $Y$, and remaining constraint variables $S$. For a known deterministic constraint, admissible activation gives a resolution map $Y = g_{z,Y}(X, S)$. The interventional law of $Y$ is therefore the pushforward of the distribution of $S$ under the map $s \mapsto g_{z,Y}(x, s)$:

$$P_Y^{\mathrm{do}_z(X=x;Y)} = \mathrm{Law}\left(g_{z,Y}(x, S)\right)$$

$$P\big(Y \in A \mid \mathrm{do}_z(X = x; Y)\big) = \int \mathbf{1}_A\left(g_{z,Y}(x, s)\right)\, dP_S(s)$$

This is a pushforward relation, not a backdoor-adjustment formula. For a continuous target, a density-level representation uses a Dirac delta or, in the stochastic case, the conditional transition kernel induced by the exogenous noise; an ordinary indicator of point equality is not a density. If the constraint contains exogenous noise, the deterministic map above is replaced by the corresponding Markov kernel and integrated over the distribution of S. The distinction separates evaluation of a known activated constraint from identification of that constraint from observational data.

### 5.5 Counterfactual Semantics

Let z = ({X, Y}, φz, Ωz, τz), and let u be a realization of the exogenous variables entering the constraint. Activating with driver variable X, intervention value x, and target Y gives the counterfactual value

$$Y_{\mathrm{do}_z(X=x;Y)}(u) = g_{z,Y}(x, u_z)$$

Keeping the same driver variable X and target Y while changing the intervention value from x to x' gives the within-law contrast

$$Y_{\mathrm{do}_z(X=x;Y)}(u) - Y_{\mathrm{do}_z(X=x';Y)}(u) = g_{z,Y}(x, u_z) - g_{z,Y}(x', u_z)$$

This measures the effect of changing the intervention value for a fixed driver-target labeling while holding the background realization fixed. Changing the driver variable itself, for example from Xi to Xj, compares distinct activated SCMs and is meaningful only after specifying how their exogenous variables and retained background state are coupled. When the driver and solved-for target both change, the two worlds likewise contain different structural assignments; their comparison is a cross-model counterfactual rather than a contrast internal to one activated SCM.

### 5.6 Evaluation and Identifiability

Two questions must be distinguished. If the constraint $\varphi_z$ is known, deterministic, locally invertible for Y, and activated admissibly, then the interventional law is evaluated directly by the pushforward in Section 5.4 provided the retained variables S are observed, fixed, or assigned a known distribution. Backdoor or front-door adjustment is not required merely to evaluate this known resolution map.

If $\varphi_z$, the distribution of S, or other directed mechanisms must instead be learned from observational data, identification additionally depends on support or positivity, consistency, correct model specification, measurement-error assumptions, and ordinary causal identification for any remaining directed paths in the activated graph. An unobserved member of a constraint is an unmeasured input, latent state, or source of structural heterogeneity; it is a confounder only when it is a common cause of the intervention variable and the target in the activated graph. Front-door, instrumental-variable, or other identification structure may then be relevant, but not solely because the constraint variable is unobserved.

### 5.7 Taxonomy

Causal zeros can arise from several possible mechanisms. Type I, equilibrium laws (PV = nRT): every variable is typically both an admissible driver and an admissible target. Type II, conservation laws (energy balance): the conserved total is often not an admissible target, so τz is a strict subset of Vz. Type III, definitional identities (for example, a quantity defined algebraically from others): direction is conventional and τz is fixed by which quantity is treated as defined. Type IV, accounting identities (Y = C + I + G): every component may be algebraically targetable, but the identity encodes no mechanism, so activation should be read as bookkeeping rather than physical causation. Type V, simultaneous systems (IS-LM): several coupled constraints share variables, and activation requires solving the coupled system for a chosen target vector, admissible only when the system is locally exactly identified. The CDE grounding in Section 6 applies directly to Type I and to dynamically generated instances of Type V; Types III and IV enter the ECM as orientation-free algebraic constraints without necessarily possessing an underlying relaxation dynamics.

### 5.8 Relation to Causal Constraints Models and Non-Causal Constraint Frameworks

The proposed formalism should be compared directly with the Causal Constraints Models (CCMs) of Blom, Bongers and Mooij[19], the constraint-augmented causal models of Beckers, Halpern and Hitchcock[55], and the Equilibrium Causal Models of Ryan and Dablander[38]. Table 1 summarizes the relation among these approaches and the principal dynamical extensions.

CCMs[19] are the closer formalism. A CCM constraint with its activation set and a causal zero with its driver/target sets share one intuition: a lawlike relation should be storable without a committed direction, with its role under intervention governed by explicit activation data. The difference is sharp and, we think, the whole of our contribution here: a CCM activation set is an admissibility test - under a given intervention a constraint is active and must hold, or it is inactive, and an inconsistent intervention simply has no solution - whereas Rule 0 additionally compiles the active constraint into a directed structural equation for a named target, yielding a predicted value and thereby a resolution function. Rule 0 can be read as answering a question CCMs pose but do not resolve: mechanically, what is the equation for the solved-for variable once the constraint is in force? The target-labeling that makes Rule 0 well posed is compatible with both the CCM construction and the labeled-equilibrium approach of Mooij, Janzing and Schölkopf[20].

The Beckers–Halpern–Hitchcock[55] framework is a more distant relative, aimed at genuinely non-causal relations such as part–whole identities, unit conversions, coordinate changes, with a disconnection operation that removes a variable from a causal equation. That is complementary rather than competing: their constraints are non-causal in a categorical, definitional sense, whereas a causal zero is a part of causal formalism but not yet oriented at specification time.

Two objections deserve direct engagement. First, Pearl[56], replying to Neuberg, argued that equilibrium constraints such as a supply-demand identity need no new primitive: one may descend to a deeper level of structural equations at which the relation is recovered as a consequence of fully oriented mechanisms. This is a real objection to the whole causal-zero program - it holds that any causal zero merely marks an under-refined model. Our reply is not to deny it but to identify the deeper level: Section 6 shows that the refinement Pearl invokes is, in the cases at hand, the differential-equation description whose equilibrium manifold the causal zero defines. Descending to a finer level and positing a causal zero are then the same move, related by Theorem 6.1, not a contradiction. The causal zero is a faithful summary of the dynamics, valid precisely in the stationary regime.

Second, Bochman and Lifschitz[26] observe that a Pearl structural equation encodes direction in its assignment syntax, distinct from the equality it satisfies. This reinforces our claim: it is exactly why $PV = nRT$, an equality, is not any single one of the assignments $P \coloneqq nRT/V, V \coloneqq nRT/P$, and so on - the causal zero stores the equality, and activation chooses the assignment. Related philosophical treatments of manipulation-relative and constraint causation by Woodward[57], Gillies[58], Cartwright[59], Weinberger[60] and Ross[61] situate this move within a broader interventionist tradition.

## 6. Causal Differential Equations: A Unified Dynamic Framework

### 6.1 Motivation: Time as the Resolver

Section 5 formalizes symmetric constraints but leaves open where they come from. The ideal gas law can be viewed as the macroscopic balance reached after microscopic momentum and energy exchange, but the pressure-relaxation equation used below is a reduced phenomenological model rather than a derivation from molecular dynamics. This motivates representing causal structure through differential equations and defining dynamically grounded causal zeros as local defining functions of equilibrium sets. For finite-propagation, well-posed state models, feedback can then be represented by temporal coupling and time unrolling; the static SCM is recovered when the dynamics relax to a unique attracting state and only that stationary regime is observed. Recent Hamiltonian Causal Models similarly place interventions and causal effects at the trajectory level and connect them to non-equilibrium thermodynamics[62].

### 6.2 Definition

***Definition 6.1 (Causal Differential Equation System)***

A Causal Differential Equation System (CDES) is a tuple

$$\mathcal{C} = (V, F, U, \mathcal{I}, \mathcal{T})$$

with system variables $V = \{X_1, \ldots, X_n\}$; dynamics

$$\frac{dX_i}{dt} = f_i(X_1, \ldots, X_n, U_i, t), \quad i = 1, \ldots, n$$

exogenous input processes $U$ with distribution $P_U$ over trajectories; admissible interventions $\mathcal{I}$; and a time domain beginning at $t = 0$ and extending indefinitely. We assume $F$ is locally Lipschitz so that solutions exist and are unique. A coordinate with $f_i \equiv 0$ is a constant input, parameter, or conserved coordinate; it is not merely slow. Slow coordinates require an explicit time-scale separation, for example $\frac{dX_i}{dt} = \varepsilon g_i$ with $0 < \varepsilon \ll 1$. Constant parameters and slow coordinates can both parameterize families of equilibria, but they play different mathematical roles and should not be conflated.

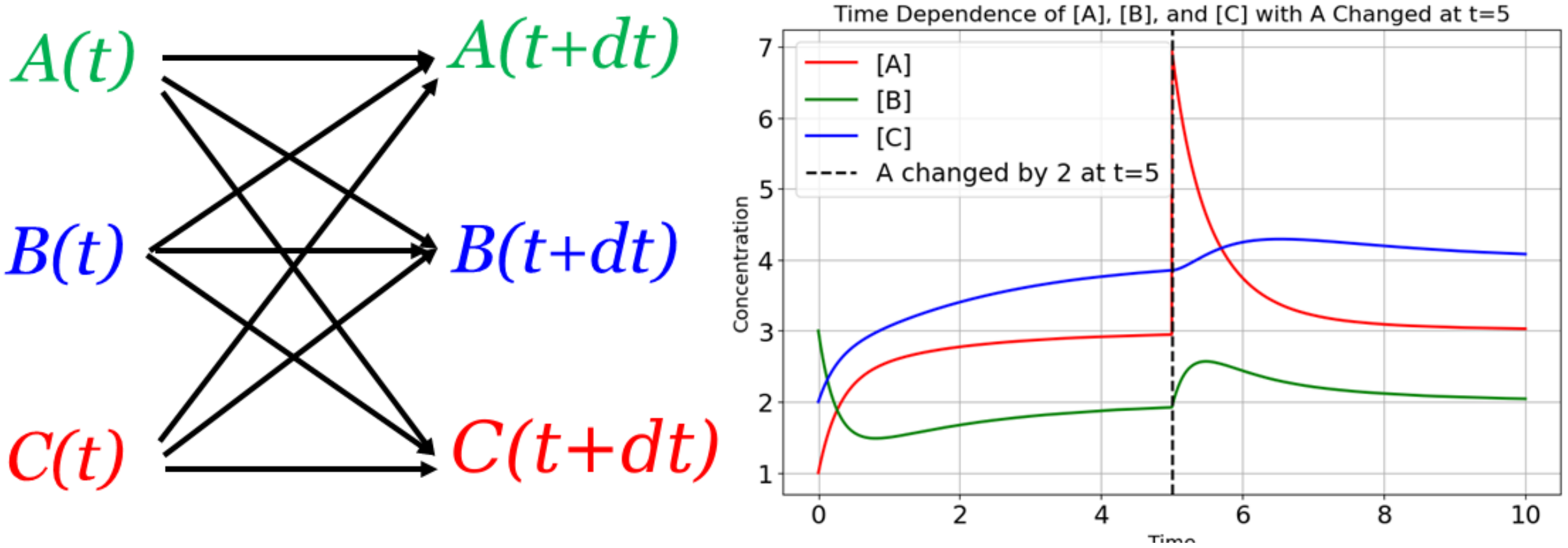

**Figure 3.** Chemical dynamics can be represented by a system of differential equations. For the finite-propagation state-space setting considered here, the appropriate causal representation is a set of causal differential equations unfolding in time rather than a literally instantaneous causal cycle.

### 6.3 Solution Regimes

***Regime 1: transient dynamics (fully causal)***

Away from equilibrium, the solution X(t) follows a trajectory whose local interaction structure is encoded by the Jacobian of F: an influence from Xj(t) to Xi(t + dt) is present when ∂fi/∂Xj is nonzero. The same-time Jacobian may contain reciprocal couplings, but for a finite-propagation model without algebraic constraints the time-unrolled graph is acyclic by construction because every edge advances time. Pearl's calculus can then be applied to the corresponding time-indexed variables, subject to the usual assumptions about the discretization and intervention semantics.

***Regime 2: equilibrium manifolds (causal zeros)***

A stationary solution satisfies

$$F(X^{*}, U) = 0$$

The set of such solutions is generally not a single point but, locally and under a constant-rank condition, a manifold whose dimension is

$$\mathcal{S} = \{X : F(X, U) = 0\} \subset \mathbb{R}^{n}, \quad dim\,\mathcal{S} = n - rank\,\partial_{X}F$$

This equilibrium manifold is the geometric object a dynamically grounded causal zero names. A scalar causal zero corresponds locally to a codimension-one component described by one defining function φz. If the equilibrium set has codimension r > 1, it requires a vector-valued constraint of rank r or, equivalently, r independent scalar causal zeros. For the gas, treating V and T as externally retained parameters, the pressure equation alone provides the reduced relaxation model

$$\frac{dP}{dt} = \alpha\left(\frac{nRT}{V} - P\right) = 0 \quad \Longrightarrow \quad PV = nRT$$

relaxes P to the surface PV = nRT, a two-dimensional family of equilibria indexed by the retained parameters (V, T). This is a controlled phenomenological relaxation of pressure toward the ideal-gas manifold, not a microscopic derivation of the ideal gas law. The rate α sets the relaxation time. The causal zero is the constraint that remains after the fast variable has equilibrated while the retained inputs are held fixed; in this sense it is an attracting equilibrium manifold rather than a unique fixed point.

***Regime 3: limit cycles and strange attractors***

When the attractor is periodic or chaotic, neither transient nor equilibrium description applies. A genuine limit cycle, i.e. an isolated attracting periodic orbit, is exhibited by the Van der Pol oscillator:

$$\frac{d^2x}{dt^2} - \mu(1 - x^2)\frac{dx}{dt} + x = 0, \quad \mu > 0$$

whose trajectories converge onto a single closed orbit regardless of initial condition. A strange attractor appears in the Lorenz system:

$$\dot{x} = \sigma(y - x), \quad \dot{y} = x(\rho - z) - y, \quad \dot{z} = xy - \beta z$$

Causal inference in this regime needs a notion of attractor-relative intervention: perturb the state off the attractor and measure the return. Because the effect depends on where in the cycle the perturbation is applied, the relevant object is a phase-dependent response, connecting to phase-response curves in mathematical biology. This regime remains open and is discussed in Section 7.4.

## 6.4 Interventions

### *6.4.1 Hard intervention (Pearl recovered)*

Pearl's equation-replacement intervention $\mathrm{do}(X_i = x)$ can be approximated within a differential equation by a high-gain forcing term:

$$\frac{dX_i}{dt} = f_i(X, U_i) + \lambda(x - X_i), \quad \lambda \to \infty$$

For finite $\lambda$ this is a soft forcing intervention. Under stability of the clamped reduced system, the ordered limit obtained by first taking the long-time limit and then the high-gain limit can recover the equilibrium response associated with equation replacement. Interchanging the gain and long-time limits requires a uniform Tikhonov-type stability condition and is not automatic. Thus the high-gain construction is an approximation to Pearl's hard intervention; exact do semantics remains the replacement of the $X_i$ equation by $X_i = x$.

### *6.4.2 Soft intervention*

At finite gain the intervention nudges rather than clamps, modeling a push toward a target that does not override the natural dynamics:

$$P^{\lambda}(y \mid \mathrm{do^{soft}}_{\lambda}(X_i = x)) = \lim_{t\to\infty} P\left(Y(t) \mid \frac{dX_i}{dt} = f_i + \lambda(x - X_i)\right)$$

As the gain runs from zero to infinity, the one-parameter family interpolates from no intervention to Pearl's hard intervention and traces an intervention-strength response.

### *6.4.3 Transient intervention*

Finite-duration interventions are naturally expressible:

$$\mathrm{do}_{[t_0,t_1]}(X_i = x): \;\; X_i(t) = x \text{ for } t \in [t_0, t_1], \qquad \text{released for } t > t_1$$

allowing the study of pulse interventions, treatment windows, and the persistence of effects after an intervention ends. Such interventions fit naturally within DSCMs and other time-varying causal models[23]. Recent work specifically distinguishes transient from lasting causal effects by examining how interventions alter equilibrium behavior and by making the measurement times part of the causal question[63].

### *6.4.4 Rate intervention*

One may also fix a rate rather than a level:

$$\mathrm{do}_v\left(\frac{dX_i}{dt} = c\right)$$

relevant to fixing growth rates in economics, absorption rates in pharmacology, or incidence rates in epidemiology, with no static analogue.

## 6.5 Causal Zeros as Equilibrium Manifolds

*Theorem 6.1 (CDE–Zero correspondence)*

Let $\mathcal{C} = (V, F, U, \mathcal{I}, \mathcal{T})$ be a CDES with a continuously differentiable vector field and a normally hyperbolic attracting equilibrium manifold $S = \{X: F(X, U) = 0\}$. Fix a driver $X_d$ and target $X_t$. Assume that, in the neighborhood of interest: (i) $S$ is represented relative to the selected target by a continuously differentiable scalar defining function $h_{z,t}(X) = 0$, or by one selected component of a full-rank vector constraint; (ii) $\partial h_{z,t}/\partial X_t \neq 0$; (iii) after clamping $X_d$, the reduced flow is stable and converges to a unique point conditional on retained coordinates $X_R$; and (iv) the high-gain and long-time limits commute.

$$h_{z,t}(X_t, X_{-t}) = 0, \qquad \frac{\partial h_{z,t}}{\partial X_t} \neq 0$$

Then $z = (V_z, h_{z,t}, \Omega_z, \tau_z)$, with $\Omega_z$ the coordinates that may be clamped and $\tau_z$ those for which the corresponding local defining function is invertible, is a causal zero in the sense of Definition 5.1. The activating intervention recovers the same local target value as the clamped dynamics taken to equilibrium; under assumption (iv), this equilibrium value is also the ordered long-time/high-gain limit:

$$X_t^{\mathrm{eq}} = g_{z,t}(x_d, X_R)$$

Proof sketch. Normal hyperbolicity gives attraction transverse to $S$. Assumption (iii) supplies the additional conditional selection of a unique point along $S$ after the driver and retained coordinates are fixed. The implicit function theorem applied to $h_{z,t}$ gives the local resolution function $g_{z,t}$; assumption (iv) equates its value with the ordered high-gain, long-time limit. If the codimension of $S$ is $r > 1$, one must use a vector defining function $h: \mathbb{R}^n \to \mathbb{R}^r$ with rank-$r$ Jacobian and activate $r$ independent constraints or an $r$-dimensional target vector. Normal hyperbolicity alone does not select a unique point along the manifold.

### 6.6 Cycles Dissolved

For finite-propagation, well-posed ODE state models without algebraic constraints, an apparent instantaneous cycle between two measured variables can be represented by coupled differential equations and an acyclic time unrolling: each influence connects a state at time t to a state at a later time. The same-time loop is then the projection of a legitimate temporal process rather than a literal instantaneous ancestry relation.

$$\text{apparent cycle} \;=\; \text{transient temporal feedback} \;+\; \text{(possibly) stationary causal zero}$$

This is a modeling claim for the stated class, not a universal prohibition on cyclic causal models. At coarser static, equilibrium, or sample-path levels, cyclic SCMs and causal SDEs can be well defined and can support sigma-separation and do-calculus when the relevant solvability conditions hold[21, 22, 24]. The CDE representation is useful when the physical time evolution itself is available and should remain explicit.

### 6.7 Counterfactuals

Counterfactual inference extends Pearl's three steps. Abduction infers the exogenous inputs from the observed trajectory,

$$P(U \mid \{X(t)\}) \propto P(\{X(t)\} \mid U)\, P(U)$$

action modifies the vector field by the counterfactual intervention,

$$f_i^{\mathrm{CF}}(X, U_i) = f_i(X, U_i) + g_{\mathrm{intervention}}(X, t)$$

and prediction integrates the modified system forward,

$$Y_{\mathrm{CF}}(u) = Y^{\mathrm{CF}}(t \to \infty \mid U = u)$$

For a system that relaxes to equilibrium this recovers Pearl's counterfactual as the equilibrium value; for transient or oscillatory systems the counterfactual is an entire trajectory, a strict generalization.

### 6.8 Pearl's SCM as a Special Case

***Proposition 6.2***

A regular continuous-valued Pearl SCM with Lipschitz structural assignments can be embedded in the relaxation CDES

$$f_i^{\mathrm{CDE}}(X, U_i, t) = f_i^{\mathrm{Pearl}}(\mathrm{PA}_i, U_i) - X_i \qquad \text{(relaxation dynamics)}$$

provided the relaxation is globally asymptotically stable, its unique fixed point is the SCM structural assignment and its graph at equilibrium is G. Within this regular class, an SCM is the special case in which the dynamics relax to an isolated stable point and only the stationary regime is observed. This proposition does not cover arbitrary discrete, discontinuous, or non-differentiable SCMs. Causal zeros appear when the equilibrium set has positive dimension; limit cycles and strange attractors arise when the attracting set is not a stable point.

## 7. Open Problems

### 7.1 Completeness of the Extended do-Calculus

The three rules are complete for DAG-structured SCMs. Whether Rules 0–3 are complete for ECMs is open. The difficulty is that Rule 0 changes the graph and, because activation is target-indexed, a single ECM query may admit several activated graphs. A completeness result would need to quantify over admissible (driver, target) labelings and show the rules capture every identifiable query.

### 7.2 Separation for Hypergraphs

*d*-separation characterizes independence in DAGs. An ECM with constraint hyperedges needs a separation criterion covering the symmetric coupling before activation, the directed structure after it, and their interaction. The *m*-separation of Richardson and Spirtes[41] and the sigma-separation of Forré and Mooij[22] are the natural starting points.

### 7.3 Stochastic CDEs

Replacing the ODEs by stochastic differential equations,

$$dX_i \;=\; f_i(X,t)\,dt \;+\; \sigma_i(X,t)\,dW_i$$

When a stationary density exists, it replaces a deterministic equilibrium surface as the object a probabilistic causal zero may summarize, with the stationary Fokker-Planck equation providing the corresponding constraint. Sokol and Hansen[50] give postintervention semantics directly for SDEs, and Lorch, Krause and Schölkopf[64] construct graph-free causal models from stationary diffusion densities. Boeken and Mooij[24] provide a complementary path-level theory for causal SDEs, establishing sigma-separation and do-calculus under solvability conditions and d-separation for additive-noise systems even in the presence of cycles. These results supply a direct bridge for extending causal zeros beyond deterministic equilibrium manifolds.

### 7.4 Attractor-Relative Interventions

For limit cycles and strange attractors, intervention must be defined relative to the attractor: perturb, then measure the return, with an effect that depends on the phase at which the perturbation is applied. A theory of phase-indexed interventional distributions and phase-shifted counterfactuals is needed, connecting to phase-response curves and to the ergodic theory of the attractor. The Van der Pol and Lorenz systems of Section 6.3 are natural test cases.

### 7.5 Learning Causal Zeros

Discovering causal zeros means distinguishing a symmetric constraint from a directed mechanism. The signature is intervention-reversal: through a causal zero, fixing the target's value back-determines the driver via the constraint, whereas through a directed mechanism, intervening on the effect leaves the cause unchanged, since surgery removes the incoming arrow. A designed sequence of interventions on different members of a candidate constraint can separate the two.

### 7.6 Applications

The framework applies directly to thermodynamics and statistical mechanics, where equilibrium laws are stationary limits of kinetics; to macroeconomics, where simultaneous systems become CDEs with equilibrium causal zeros; to systems biology, where directed regulation coexists with equilibrium binding constraints; and to control theory, where the soft and rate interventions of Section 6.4 correspond to proportional and derivative control.

## 8. Discussion

The paper advances two extensions unified by one observation: not every scientific relationship is directional, and not every dynamical process is faithfully a static acyclic graph. The first extension, causal zeros within an Extended Causal Model, gives a language for symmetric constraints that acquire direction only under intervention. The activation operator compiles such a constraint into a structural equation by naming a driver and a target, after which the model is an ordinary SCM. The essential correction over the initial proposal is that activation must name a target: one scalar constraint determines one variable, and pretending otherwise reproduces exactly the ambiguity that labeled equilibrium equations were introduced to remove.

The second extension, Causal Differential Equations, grounds the static picture in dynamics. Dynamically grounded causal zeros are local defining functions of attracting equilibrium manifolds; a regular continuous SCM is the case of relaxation to an isolated stable point observed only at rest. This also answers Pearl's own objection to positing new primitives: the deeper level his reply invokes is the differential-equation description itself, and the causal zero is its faithful equilibrium summary.

Together the extensions address the two structural limits of the static framework - the absence of undirected symmetric relations and the exclusion of feedback - while conceding that the first extension refines, rather than replaces, the Causal Constraints Models[19] it most resembles. The central shift is to treat causal structure as partly constituted by intervention: a causal zero has no direction until an intervention selects a driver and a target; a CDES has no static DAG until time is unrolled or an equilibrium abstraction is selected. This is congenial to, though more thoroughgoing than, Woodward's[57] interventionism. The temporal claim is restricted to finite-propagation, well-posed state models; cyclic SCM and SDE frameworks remain valid under their own solvability conditions.

## 9. Conclusion

Causal zeros are symmetric constraint equations such as equilibrium laws, conservation laws, accounting identities, and simultaneous systems carrying no intrinsic direction, stored as constraint hyperedges in an Extended Causal Model and compiled into directed mechanisms by an activation operator that names both a driver and a target. Activation yields an ordinary DAG-based SCM only when local solvability, compatibility, and graph-admissibility conditions hold. Causal Differential Equations ground the dynamically generated part of the construction in continuous time: transient solutions are time-unrolled causal processes, and scalar causal zeros are local defining functions of codimension-one attracting equilibrium manifolds, with regular continuous SCMs recovered in the isolated-fixed-point limit. For the finite-propagation state-space class considered here, apparent same-time cycles are represented as temporal feedback. The framework remains a beginning: completeness, hypergraph separation, stochastic causal zeros, attractor-relative intervention, and discovery remain open. Its contribution over the closest prior work is narrow but real - an orientation operator, correctly target-labeled and explicitly admissibility-conditioned, sitting on top of the constraint-model and dynamical-SCM programs it builds on.

**Acknowledgement:**

This study was conducted by the author as a curiosity-driven exploration begun in 2020 and was formalized with Fable 5.

## References

(1) Pearl, J. *Causality: Models, Reasoning, and Inference*; Cambridge University Press, 2009.
(2) Spirtes, P.; Glymour, C.; Scheines, R. *Causation, Prediction, and Search*; MIT Press, 2000.
(3) Runge, J.; Bathiany, S.; Bollt, E.; Camps-Valls, G.; Coumou, D.; Deyle, E.; Glymour, C.; Kretschmer, M.; Mahecha, M. D.; Muñoz-Marí, J.; van Nes, E. H.; Peters, J.; Quax, R.; Reichstein, M.; Scheffer, M.; Schölkopf, B.; Spirtes, P.; Sugihara, G.; Sun, J.; Zhang, K.; Zscheischler, J. Inferring causation from time series in Earth system sciences. *Nature Communications* **2019**, *10*, 2553.
(4) Runge, J.; Gerhardus, A.; Varando, G.; Eyring, V.; Camps-Valls, G. Causal inference for time series. *Nature Reviews Earth & Environment* **2023**, *4*, 487-505.
(5) Camps-Valls, G.; Gerhardus, A.; Ninad, U.; Varando, G.; Martius, G.; Balaguer-Ballester, E.; Vinuesa, R.; Diaz, E.; Zanna, L.; Runge, J. Discovering causal relations and equations from data. *Physics Reports* **2023**, *1044*, 1-68.
(6) Ziatdinov, M.; Nelson, C. T.; Zhang, X.; Vasudevan, R. K.; Eliseev, E.; Morozovska, A. N.; Takeuchi, I.; Kalinin, S. V. Causal analysis of competing atomistic mechanisms in ferroelectric materials from high-resolution scanning transmission electron microscopy data. *npj Computational Materials* **2020**, *6*, 127.
(7) Nelson, C. T.; Vasudevan, R. K.; Zhang, X.; Ziatdinov, M.; Eliseev, E. A.; Takeuchi, I.; Morozovska, A. N.; Kalinin, S. V. Exploring physics of ferroelectric domain walls via Bayesian analysis of atomically resolved STEM data. *Nature Communications* **2020**, *11*, 6361.
(8) Barakati, K.; Nelson, C. T.; Morozovska, A. N.; Ziatdinov, M. A.; Eliseev, E. A.; Zhang, X.; Takeuchi, I.; Kalinin, S. V. Mapping causal patterns in crystalline solids. *APL Machine Learning* **2025**, *3*, 036117. DOI: 10.1063/5.0284465.
(9) Ziatdinov, M.; Creange, N.; Zhang, X.; Morozovska, A. N.; Eliseev, E.; Vasudevan, R. K.; Takeuchi, I.; Nelson, C. T.; Kalinin, S. V. Predictability as a probe of manifest and latent physics: The case of atomic scale structural, chemical, and polarization behaviors in multiferroic Sm-doped BiFeO3. *Applied Physics Reviews* **2021**, *8*, 011403.
(10) Shimizu, S.; Hoyer, P. O.; Hyvärinen, A.; Kerminen, A. A linear non-Gaussian acyclic model for causal discovery. *Journal of Machine Learning Research* **2006**, *7*, 2003-2030.
(11) Liu, Y.; Ziatdinov, M.; Kalinin, S. V. Exploring causal physical mechanisms via non-Gaussian linear models and deep kernel learning: Applications for ferroelectric domain structures. *ACS Nano* **2022**, *16*, 1250-1259.
(12) Kalinin, S. V.; Ghosh, A.; Vasudevan, R.; Ziatdinov, M. From atomically resolved imaging to generative and causal models. *Nature Physics* **2022**, *18*, 1152-1160.
(13) Kalinin, S. V.; Vasudevan, R.; Liu, Y.; Ghosh, A.; Roccapriore, K.; Ziatdinov, M. Probe microscopy is all you need. *Machine Learning: Science and Technology* **2023**, *4*, 023001.
(14) Barakati, K.; Molak, A.; Nelson, C. T.; Zhang, X.; Takeuchi, I.; Kalinin, S. V. Causal discovery from data assisted by large language models. *Applied Physics Letters* **2025**, *127*, 121904.
(15) Ghosh, A.; Palanichamy, G.; Trujillo, D. P.; Shaikh, M.; Ghosh, S. Insights into cation ordering of double perovskite oxides from machine learning and causal relations. *Chemistry of Materials* **2022**, *34*, 7563-7578.
(16) Gocho, K.; Hamaie, M.; Tanibata, N.; Takeda, H.; Nakayama, M.; Karasuyama, M.; Kobayashi, R. Causal analysis of factors for Li ionic conductivity in olivine-type LiMXO4 materials using LiNGAM. *The Journal of Physical Chemistry C* **2025**, *129*, 1035-1043.
(17) Ting, J. Y. C.; Barnard, A. S. Data-driven causal inference of process-structure relationships in nanocatalysis. *Current Opinion in Chemical Engineering* **2022**, *36*, 100818.

(18) Ghosh, A. Towards physics-informed explainable machine learning and causal models for materials research. *Computational Materials Science* **2024**, *233*, 112740.
(19) Blom, T.; Bongers, S.; Mooij, J. M. Beyond structural causal models: Causal constraints models. In *Proceedings of the 35th Conference on Uncertainty in Artificial Intelligence (UAI), PMLR 115*, 2020; pp 585-594.
(20) Mooij, J. M.; Janzing, D.; Schölkopf, B. From ordinary differential equations to structural causal models: The deterministic case. In *Proceedings of the 29th Conference on Uncertainty in Artificial Intelligence (UAI)*, 2013; pp 440-448.
(21) Bongers, S.; Forré, P.; Peters, J.; Mooij, J. M. Foundations of structural causal models with cycles and latent variables. *Annals of Statistics* **2021**, *49*, 2885-2915.
(22) Forré, P.; Mooij, J. M. Causal calculus in the presence of cycles, latent confounders and selection bias. In *Proceedings of the 35th Conference on Uncertainty in Artificial Intelligence (UAI), PMLR 115*, 2019; pp 71-80.
(23) Boeken, P.; Mooij, J. M. Dynamic structural causal models. In *arXiv*, 2024.
(24) Boeken, P.; Mooij, J. M. Causal Graphs, Markov Properties and Do-calculus for Stochastic Differential Equations. In *arXiv*, 2026.
(25) Pearl, J. Causal diagrams for empirical research. *Biometrika* **1995**, *82* (4), 669-688.
(26) Bochman, A.; Lifschitz, V. Pearl's causality in a logical setting. In *Proceedings of the 29th AAAI Conference on Artificial Intelligence*, 2015; pp 1446-1452.
(27) Shpitser, I.; Pearl, J. Identification of joint interventional distributions in recursive semi-Markovian causal models. In *Proceedings of the 21st National Conference on Artificial Intelligence (AAAI)*, 2006; pp 1219-1226.
(28) Huang, Y.; Valtorta, M. Pearl's calculus of intervention is complete. In *Proceedings of the 22nd Conference on Uncertainty in Artificial Intelligence (UAI)*, 2006; pp 217-224.
(29) Pearl, J.; Mackenzie, D. *The Book of Why: The New Science of Cause and Effect*; Basic Books, 2018.
(30) Simon, H. A. Causal ordering and identifiability. In *Studies in Econometric Method*, Wiley, 1953; pp 49-74.
(31) Iwasaki, Y.; Simon, H. A. Causality and model abstraction. *Artificial Intelligence* **1994**, *67* (1), 143-194.
(32) Spirtes, P. Directed cyclic graphical representations of feedback models. In *Proceedings of the 11th Conference on Uncertainty in Artificial Intelligence (UAI)*, 1995; pp 491-498.
(33) Strotz, R. H.; Wold, H. O. A. Recursive versus nonrecursive systems: An attempt at synthesis. *Econometrica* **1960**, *28* (2), 417-427.
(34) Mooij, J. M.; Janzing, D.; Heskes, T.; Schölkopf, B. On causal discovery with cyclic additive noise models. In *Advances in Neural Information Processing Systems (NeurIPS) 24*, 2011; pp 639-647.
(35) Mooij, J. M.; Heskes, T. Cyclic causal discovery from continuous equilibrium data. In *Proceedings of the 29th Conference on Uncertainty in Artificial Intelligence (UAI)*, 2013; pp 431-439.
(36) Bongers, S.; Blom, T.; Mooij, J. M. Causal modeling of dynamical systems. In *arXiv*, 2018.
(37) Rubenstein, P. K.; Bongers, S.; Schölkopf, B.; Mooij, J. M. From deterministic ODEs to dynamic structural causal models. In *Proceedings of the 34th Conference on Uncertainty in Artificial Intelligence (UAI)*, 2018; pp 114-123.

(38) Ryan, O.; Dablander, F. Equilibrium Causal Models: Connecting Dynamical Systems Modeling and Cross-Sectional Data Analysis. *Multivariate Behavioral Research* **2025**, *60* (6), 1116-1150. DOI: 10.1080/00273171.2025.2522733.
(39) Forré, P.; Mooij, J. M. Constraint-based causal discovery for non-linear structural causal models with cycles and latent confounders. In *Proceedings of the 34th Conference on Uncertainty in Artificial Intelligence (UAI)*, 2018; pp 269-278.
(40) Mooij, J. M.; Claassen, T. Constraint-based causal discovery using partial ancestral graphs in the presence of cycles. In *Proceedings of the 36th Conference on Uncertainty in Artificial Intelligence (UAI), PMLR 124*, 2020; pp 1159-1168.
(41) Richardson, T. S.; Spirtes, P. Ancestral graph Markov models. *Annals of Statistics* **2002**, *30* (4), 962-1030.
(42) Richardson, T. S. Markov properties for acyclic directed mixed graphs. *Scandinavian Journal of Statistics* **2003**, *30* (1), 145-157.
(43) Dean, T.; Kanazawa, K. A model for reasoning about persistence and causation. *Computational Intelligence* **1989**, *5* (3), 142-150.
(44) Murphy, K. P. Dynamic Bayesian Networks: Representation, Inference and Learning. University of California, Berkeley, 2002.
(45) Granger, C. W. J. Investigating causal relations by econometric models and cross-spectral methods. *Econometrica* **1969**, *37* (3), 424-438.
(46) Haavelmo, T. The statistical implications of a system of simultaneous equations. *Econometrica* **1943**, *11* (1), 1-12.
(47) *Statistical Inference in Dynamic Economic Models*; Wiley, 1950.
(48) Hyttinen, A.; Eberhardt, F.; Hoyer, P. O. Learning linear cyclic causal models with latent variables. *Journal of Machine Learning Research* **2012**, *13*, 3387-3439.
(49) Hyttinen, A.; Eberhardt, F.; Järvisalo, M. Constraint-based causal discovery: Conflict resolution with answer set programming. In *Proceedings of the 30th Conference on Uncertainty in Artificial Intelligence (UAI)*, 2014; pp 340-349.
(50) Sokol, A.; Hansen, N. R. Causal interpretation of stochastic differential equations. *Electronic Journal of Probability* **2014**, *19* (100), 1-24.
(51) Eichler, M.; Didelez, V. On Granger causality and the effect of interventions in time series. *Lifetime Data Analysis* **2010**, *16* (1), 3-32.
(52) Eichler, M. Graphical modelling of multivariate time series. *Probability Theory and Related Fields* **2012**, *153* (1-2), 233-268.
(53) White, H.; Chalak, K.; Lu, X. Linking Granger causality and the Pearl causal model with settable systems. *JMLR Workshop and Conference Proceedings* **2011**, *12*, 1-29.
(54) Røysland, K.; Ryalen, P. C.; Nygård, M.; Didelez, V. Graphical criteria for the identification of marginal causal effects in continuous-time survival and event-history analyses. *Journal of the Royal Statistical Society Series B: Statistical Methodology* **2025**, *87* (1), 74-97. DOI: 10.1093/jrsssb/qkae056.
(55) Beckers, S.; Halpern, J. Y.; Hitchcock, C. Causal models with constraints. In *Proceedings of the Second Conference on Causal Learning and Reasoning (CLeaR), PMLR 213*, 2023; pp 866-879.
(56) Pearl, J. Reply to Neuberg's review of Causality. *Econometric Theory* **2003**, *19* (4), 686-689.
(57) Woodward, J. *Making Things Happen: A Theory of Causal Explanation*; Oxford University Press, 2003.

(58) Gillies, D. An action-related theory of causality. *British Journal for the Philosophy of Science* **2005**, *56* (4), 823-842.
(59) Cartwright, N. *How the Laws of Physics Lie*; Oxford University Press, 1983.
(60) Weinberger, N. Intervening and Letting Go: On the Adequacy of Equilibrium Causal Models. *Erkenntnis* **2023**, *88* (6), 2467-2491. DOI: 10.1007/s10670-021-00463-0.
(61) Ross, L. N. Causal constraints in the life and social sciences. *Philosophy of Science* **2024**, *91* (5), 1068-1077. DOI: 10.1017/psa.2023.165.
(62) Rancati, D.; Welling, M.; Locatello, F. Reconciling Causality and Non-Equilibrium Thermodynamics with Hamiltonian Causal Models. In *arXiv*, 2026.
(63) Steele, R.; Weinberger, N.; Baker, T.; Shrier, I. Modifying causal models to distinguish between transient and lasting causal effects. In *arXiv*, 2026.
(64) Lorch, L.; Krause, A.; Schölkopf, B. Causal modeling with stationary diffusions. In *Proceedings of the 27th International Conference on Artificial Intelligence and Statistics (AISTATS), PMLR 238*, 2024; pp 1927-1935.